\pdfoutput=1

\documentclass[11pt]{article}

\usepackage[]{ACL2023}

\usepackage{times}
\usepackage{latexsym}
\usepackage{hyperref}
\usepackage{url}
\usepackage{graphicx}
\usepackage{natbib}
\usepackage{subcaption}
\usepackage{float}
\usepackage{booktabs}
\usepackage{colortbl}
\usepackage{makecell}

\usepackage[T1]{fontenc}

\usepackage[utf8]{inputenc}

\usepackage{microtype}

\usepackage{inconsolata}

\newcommand{\ourmodel}[1]{OPT-\textit{#1}}

\newcommand{\datasetname}[1]{#1}
\newcommand{\superni}[0]{S{\small UPER}-N{\small ATURAL}I{\small NSTRUCTIONS} }
\newcommand{\supernishort}[0]{S{\small UP}-N{\small AT}I{\small NST} }

\title{\ourmodel{R}: Exploring the Role of Explanations in Finetuning and Prompting for Reasoning Skills of Large Language Models}
\author{Badr AlKhamissi \quad Siddharth Verma \quad Ping Yu \quad Zhijing Jin \\
{\bf Asli Celikyilmaz \quad Mona Diab}
\\
{\normalfont Meta AI}}

\begin{document}

\maketitle

\begin{abstract}

In this paper, we conduct a thorough investigation into the reasoning capabilities of Large Language Models (LLMs), focusing specifically on the Open Pretrained Transformers (OPT) models as a representative of such models. Our study entails finetuning three different sizes of OPT on a carefully curated reasoning corpus, resulting in two sets of finetuned models: \ourmodel{R}, finetuned without explanations, and \ourmodel{RE}, finetuned with explanations. We then evaluate all models on $57$ out-of-domain tasks drawn from the \superni benchmark, covering $26$ distinct reasoning skills, utilizing three prompting techniques. Through a comprehensive grid of 27 configurations and 6,156 test evaluations, we investigate the dimensions of finetuning, prompting, and scale to understand the role of explanations on different reasoning skills. Our findings reveal that having explanations in the fewshot exemplar has no significant impact on the model's performance when the model is finetuned, while positively affecting the non-finetuned counterpart. Moreover, we observe a slight yet consistent increase in classification accuracy as we incorporate explanations during prompting and finetuning, respectively. Finally, we offer insights on which skills benefit the most from incorporating explanations during finetuning and prompting, such as \textit{Numerical} ($+20.4$\%) and \textit{Analogical} ($+13.9$\%) reasoning, as well as skills that exhibit negligible or negative effects.

\end{abstract}

\section{Introduction}


Recently, there has been a surge in the release of Large Language Models (LLMs) by both industrial and academic institutions. These models vary from open-source releases such as OPT \cite{opt} and LLAMA \cite{Touvron2023LLaMAOA} to closed-source ones like GPT-3 \cite{gpt3} and PALM \cite{palm2022}. In addition, researchers have developed models that are finetuned on top of these foundational models to better follow instructions, such as OPT-IML \cite{iyer2022opt} and Alpaca \cite{alpaca}. Despite the remarkable progress in LLMs' performance in Natural Language Processing (NLP) tasks, reasoning remains a challenging area. For example, prior work have shown that LLMs struggle with commonsense reasoning \cite{west2022symbolic} and arithmetic reasoning \cite{hendrycksmath2021} to name a few.

\begin{figure}[t!]
    \centering
    \includegraphics[width=1.\linewidth]{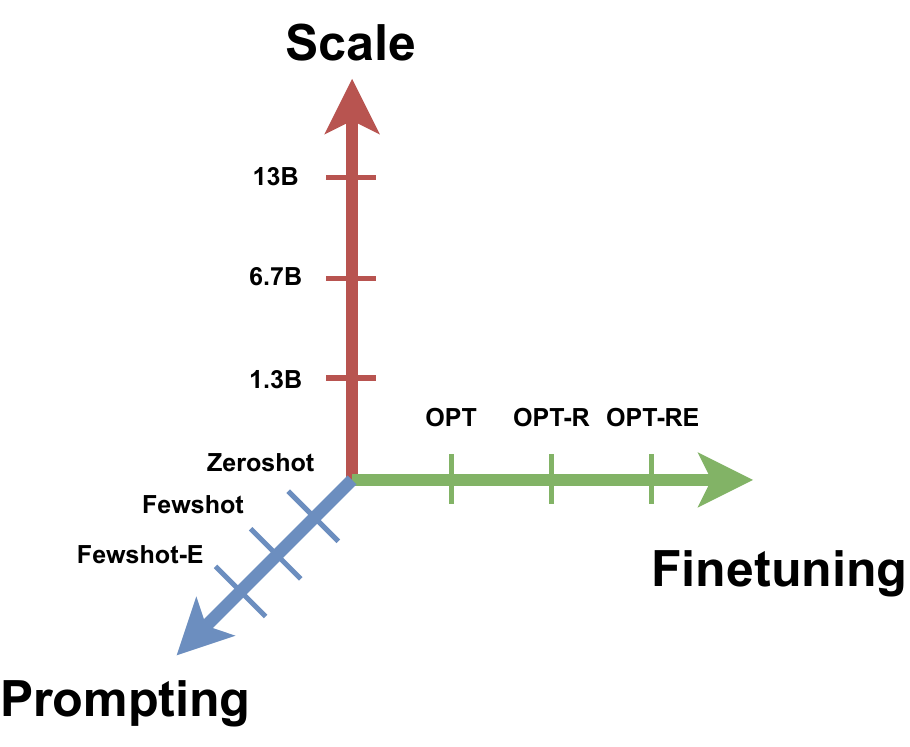}
    \caption{\textbf{Three-Dimensional Grid of Fine-Tuning, Prompting, and Scale.} Each dimension is represented as an axis, with  three  levels for each of finetuning, prompting, and scale plotted on each axis. The resulting grid consists of 27 different combinations evaluated on various reasoning tasks. It should be noted that there is a hidden dimension, the scoring function, comprising four components. This results in a comprehensive total of 6,156 evaluations.}
    \label{fig:main-dims}
\end{figure}

Recent efforts have attempted to improve the reasoning performance of LLMs by decomposing answers into step-by-step reasoning chains using in-context learning \cite{wei2022chain, kojima2022large} or during finetuning \cite{chung2022scaling, wei2021finetuned}. While these approaches have shown some improvement on benchmarks such as GSM8K \cite{Cobbe2021TrainingVT}, it is not clear how those explanations affect finetuning, prompting, or their combination. Concurrent work has investigated the generalization capability of such models to reasoning skills beyond those encountered during finetuning \cite{Yu2022ALERTAL}, but a comprehensive evaluation of the role of explanation during finetuning and prompting with respect to reasoning skills is still lacking. 

In this paper, we aim to address this gap. We investigate OPT \cite{opt} as a representative of such models and utilize it as our base model. Through finetuning OPT on a collection of carefully curated open-source reasoning datasets that come with explanations for each instance, we evaluate its performance on $57$ tasks drawn from the \superni benchmark \cite{niv2}, covering $26$ different reasoning skills. Our experiments are structured around three key dimensions: finetuning, prompting, and scale, each of which is comprised of three distinct components (See Figure \ref{fig:main-dims}). \textbf{Finetuning}: (1) a (vanilla) unfinetuned OPT model; (2) A finetuned OPT model without explanations (\ourmodel{R}); and, (3) A finetuned OPT model with explanations (\ourmodel{RE}). \textbf{Prompting}: (1) zero-shot prompting; (2) Fewshot prompting without explanations; and, (3) Fewshot prompting with explanations. Finally, \textbf{Scale}: (1) $1.3$B; (2) $6.7$B; and, (3) $13$B. Accordingly, we create grid of 27 different components, providing a detailed analysis measuring the impact of explanations during finetuning and inference across different model scales.

\begin{figure}[t!]
    \centering
    \includegraphics[width=1.\linewidth]{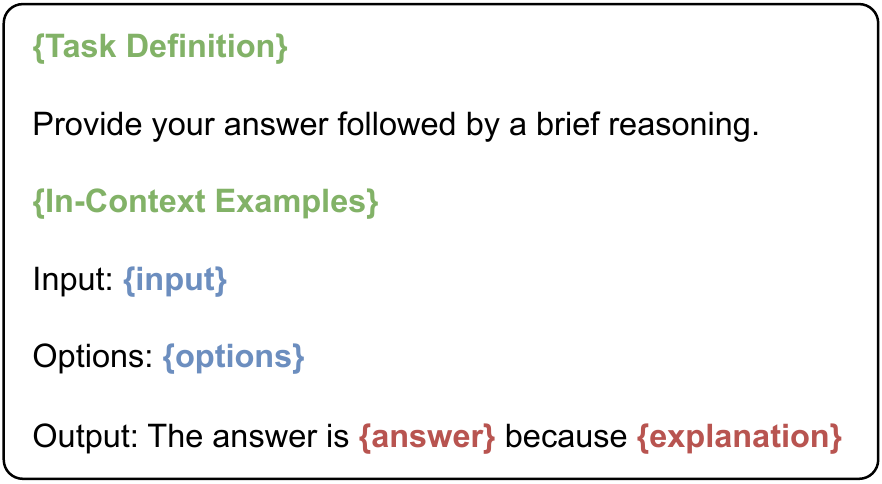}
    \caption{Template used during both training and inference. The model is tasked with predicting the answer followed by the explanation.}
    \label{fig:template}
\end{figure}

Our findings reveals that finetuning on reasoning datasets leads to statistically significant improvements in seven reasoning skills, including \textit{Numerical}, \textit{Analogical} and \textit{Reasoning on Objects}, with \textit{Physical}, \textit{Counting} and \textit{Textual Entailment} showing a significant effect only for the \ourmodel{RE} model, across both fewshot prompting conditions and model sizes, as compared to the vanilla OPT model (see Table \ref{tab:results-opt-r-sig}). However, we also find that this approach significantly hinders the performance of three other reasoning skills (see Table \ref{tab:results-opt-sig}). We also investigate the impact of incorporating explanations during fewshot prompting and find that it does not have a significant impact on the performance of the finetuned models, as measured by the variance in the difference between both prompting methods across reasoning skills for each model. However, we notice that it has a more noticeable effect on the performance of the vanilla OPT model, as shown in Table \ref{tab:results-std-prompt}. Additionally, we observe a consistent increase in the average performance across all tasks from Fewshot to Fewshot-E, as well as from OPT to \ourmodel{R} to \ourmodel{RE} models, indicating that explanations do have a small effect on performance during both finetuning and prompting. Finally, Table \ref{tab:results-color} presents a summary of the results, indicating which reasoning skills demonstrate improvement due to the incorporation of explanations during either finetuning or prompting, which skills show a negative effect, and which skills have negligible effects regarding explanations.

\section{\ourmodel{R}: Finetuning on Reasoning Skills}

\subsection{Reasoning Datasets with Explanations}
\label{sec:training-dataset}

\begin{figure}[h!]
    \centering
    \includegraphics[width=1.\linewidth]{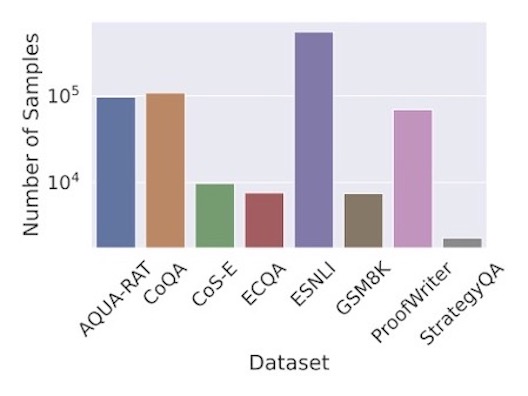}
    \caption{Number of samples in each dataset of the training corpus. Y-axis in log scale.}
    \label{fig:training_corpus}
\end{figure}

The finetuning corpus utilized to refine OPT is composed of various reasoning datasets, each of which includes a corresponding explanation or rationale for the answer. These rationales may consist of a sequence of smaller steps (i.e. chain-of-thought) or a free-form text that elucidates the reasoning behind the answer. As shown in Figure \ref{fig:template}, we employ a uniform template for all tasks during the training process. The input to the model begins with a task definition, followed by an instruction to provide an answer followed by a brief reasoning. Next, we extract two random in-context examples uniformly from the training set that remain constant throughout training for each instance. The input for the current training instance is then presented in a format specific to each task. The options for the answer are then included in the input, but not in the in-context examples (see Appendix \ref{app:finetuning-ds} for further details on task-specific definitions and options). The options are pre-shuffled for each training instance. The model is finally provided with the answer prefix, \texttt{"Output: The answer is"}, and is tasked to predict the answer, followed by an explanation if \ourmodel{RE} is being finetuned. Similarly, the in-context examples only comprise an explanation when training \ourmodel{RE}.

Below is a brief description of each dataset used during finetuning. See Figure \ref{fig:training_corpus} for the relative size of each dataset.

\paragraph{AQUA-RAT} The Algebra Question Answering with Rationales dataset \cite{ling-etal-2017-program} rendering the task of solving algebraic word problems more feasible by dividing the problem into a series of smaller steps. They create a $100$k-sample dataset that contains questions, answers and rationales in natural language and human-readable mathematical expressions that can be used to derive the final answer.

\paragraph{CoQA} The Conversational Question Answering dataset \citet{reddy-etal-2019-coqa}. It consists of $127$k questions and answers, compiled from $8$k conversations about passages from seven different domains. Given a passage that contains a conversation, the model is tasked with answering a question by highlighting the corresponding evidence from the passage.

\paragraph{CoS-E} The Common Sense Explanations dataset \citet{Rajani2019ExplainYL} to induce language models with commonsense reasoning. In this dataset, the model is given a question and a set of choices and is tasked with selecting one of the provided choices along with providing an explanation in natural language as to why that choice is correct. 

\paragraph{ECQA} The Explanations for Commonsense Question Answering dataset \citet{aggarwal-etal-2021-explanations}. It is similar to \datasetname{CoS-E} since it requires the model to choose one of the provided options to answer the given question, and also provide an explanation.

\paragraph{ESNLI} The Stanford Natural Language Inference dataset with Explanations \citet{esnli} to train models to provide interpretable and robust explanations for their decisions. The authors extend the SNLI dataset \cite{bowman2015large} with human-annotated explanations. Similar to any NLI task, the model is given a premise and hypothesis and the task is to determine whether the hypothesis sentence entails, contradicts, or is neutral with respect to the given premise.

\paragraph{GSM8K} The Grade School Math dataset \citet{Cobbe2021TrainingVT} to train models to better perform multi-step mathematical reasoning. It consists of $8.5$k linguistically diverse grade school math word problems. Therefore, the task for the model is to answer the question by performing a series of arithmetic operations to obtain a final answer, while explaining it's reasoning steps.

\paragraph{ProofWriter} The ProofWriter dataset \citet{tafjord-etal-2021-proofwriter} to generate both the implications of a theory from the RuleTaker dataset \cite{Clark2020TransformersAS} and the natural language proofs that support them. Specifically, given a sequence of facts and rules, the model is tasked with answering a question using ``Yes'', ``No'', or ``Unknown'' and provide the reasoning path by referring to the provided facts and rules. We consider the open-world assumption subset of RuleTaker with questions that requires reasoning up to a depth of $5$.

\begin{table*}[ht!]
    \centering
\begin{tabular}{ll}
\toprule
                                   \textbf{Reasoning Skill} &                     \textbf{Task IDs} \\
\midrule
                               Abductive Reasoning &                     task854 \\
                              Analogical Reasoning &          task1287, task1288 \\
                                Argument Reasoning &                     task514 \\
                                  Causal Reasoning &                    task1393 \\
                             Commonsense Reasoning &   task279, task156, task295 \\
Commonsense Reasoning $\rightarrow$ Numerical Commonsense ... &                    task1403 \\
       Commonsense Reasoning $\rightarrow$ Physical Reasoning &                     task084 \\
        Commonsense Reasoning $\rightarrow$ Social Situations &  task580, task937, task1606 \\
        Commonsense Reasoning $\rightarrow$ Spatial Reasoning &            task082, task083 \\
                               Deductive Reasoning &  task221, task1568, task220 \\
                                            Ethics &   task667, task724, task723 \\
                             Grammatical Reasoning & task1712, task052, task1559 \\
                                 Logical Reasoning &   task717, task211, task268 \\
       Logical Reasoning $\rightarrow$ Reasoning with Symbols &            task923, task935 \\
                           Mathematics $\rightarrow$ Counting &            task523, task155 \\
                                Multihop Reasoning &           task1297, task056 \\
                               Numerical Reasoning &           task621, task1333 \\
                              Reasoning on Objects &          task1583, task1584 \\
                  Reasoning on Social Interactions &   task609, task881, task875 \\
                              Reasoning on Strings &                    task1189 \\
                              Relational Reasoning & task1380, task472, task1505 \\
                              Scientific Reasoning &  task1431, task228, task714 \\
                                Temporal Reasoning &  task018, task1549, task383 \\
                                Textual Entailment &   task738, task890, task463 \\
        Textual Entailment $\rightarrow$ Analogical Reasoning &                    task1347 \\
         Textual Entailment $\rightarrow$ Deductive Reasoning & task1612, task534, task1366 \\
\bottomrule
\end{tabular}
\caption{Evaluation tasks from \supernishort \cite{niv2} used for each reasoning skill.}
\label{tab:rskill-test}
\end{table*}

\paragraph{StrategyQA} The Strategy Question Answering dataset \citet{Geva2021DidAU} to improve multi-hop reasoning for questions where the required reasoning steps are implicit in the question. Therefore, the task of the model is to answer the question using ``Yes'' or ``No'' then provide a strategy that explains the answer by decomposing it into a number of steps.

\subsection{Finetuning Procedures}

\paragraph{OPT}

The Open Pretrained Transformers (OPT) models are a suite of decoder-only pre-trained transformers ranging from $125$M to $175$B parameters released by \citet{opt}. In this work, we use three OPT models with sizes of $1.3$B, $6.7$B and $13$B. The details of each model architecture, pre-training corpus and training configuration (e.g. weight initialization, optimizer, tokenizer, hyperparameters, etc.) can be found in \citet{opt}.


\paragraph{Implementation Details}

To finetune the selected models, we utilized the metaseq\footnote{\url{https://github.com/facebookresearch/metaseq}} implementation since it enables higher training efficiency compared to other codebases \cite{opt}. Each model is finetuned twice for $10$ epochs, once with explanations and once without (i.e. \ourmodel{RE} vs \ourmodel{R}, respectively). Models are evaluated at the end of each epoch on a chosen set of \superni validation tasks, and the checkpoint with the best performance is selected for evaluation on the testing tasks. The loss is calculated only on the tokens the model is tasked to predict during inference, and not the full input, what is referred to as label-loss in \cite{iyer2022opt}. The samples across all datasets are shuffled during training. Further, the model is provided with two in-context examples during finetuning in addition to the task definition to match inference time following \cite{niv2}.






\section{Evaluating the Models}

\subsection{\superni Tasks}
\label{sec:niv2-tasks}

\begin{figure*}[ht!]
    \centering
    \includegraphics[width=1.\linewidth]{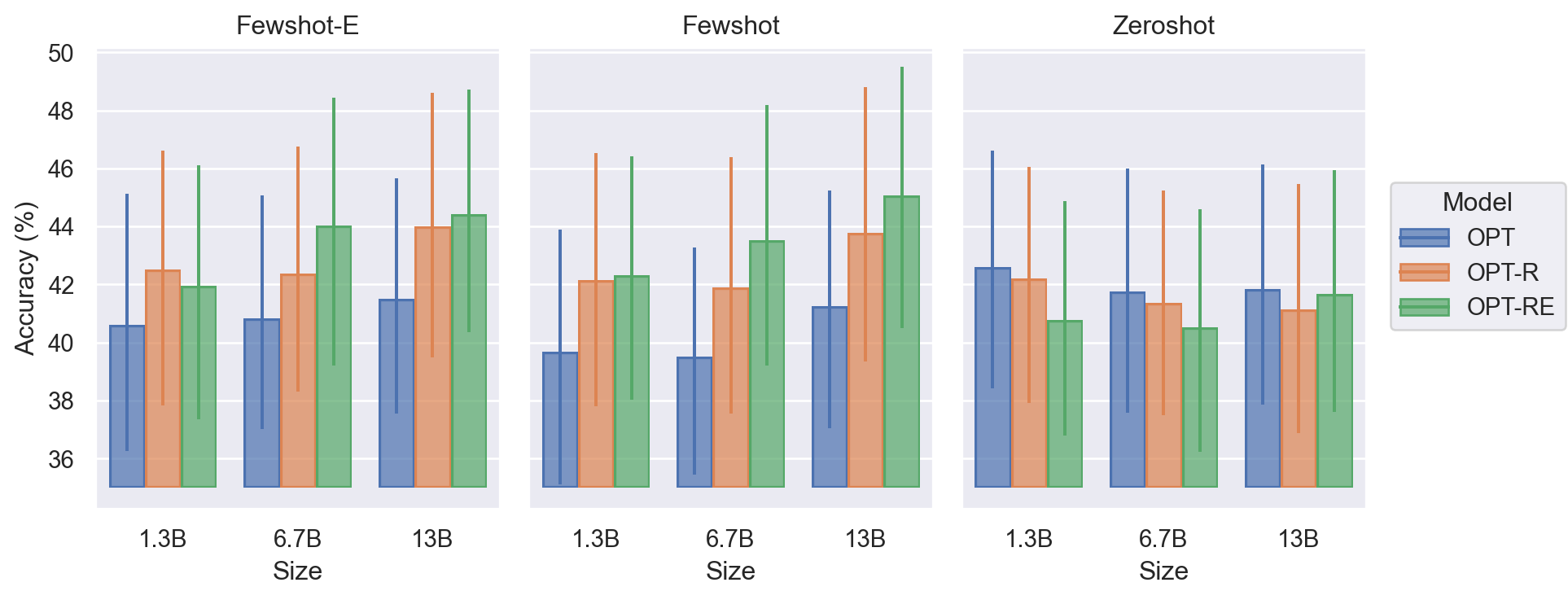}
    \caption{Results achieved across all tasks as a function of the three primary dimensions analyzed in this study: Finetuning, Prompting and Scale.}
    \label{fig:result_main}
\end{figure*}

In this study, we focus on a subset of the \superni benchmark version 2.6\footnote{We downloaded the data from \url{https://github.com/allenai/natural-instructions/tree/v2.6}.} (\supernishort for short) proposed by \citet{niv2}, which comprises 1,616 varied NLP tasks and includes meta-labels for each task, such as task type, domain and more importantly for this work: the underlying reasoning skills. Specifically, we select a subset of tasks that satisfy two key criteria: (i) the task focuses on a single reasoning skill, enabling us to evaluate a specific atomic skill, and (ii) the task can be tested using classification mode, as detailed in Section \ref{sec:eval-method}. Note that there is no data contamination between finetuning data and the evaluation benchmark.

\paragraph{Benchmark Splits} Following the task selection process, we apply a random sampling technique to ensure diversity within the testing set. Specifically, we select a maximum of three tasks from each reasoning skill, and allocate any remaining tasks to the validation set. Notably, this approach enables us to obtain a representative sample of the selected reasoning skills for testing, while also ensuring that our model's performance is not influenced by a particular subset of tasks. Table \ref{tab:rskill-test} shows the complete list of tasks used for evaluating our finetuned models for each reasoning skill.

\subsection{Evaluation Setup}
\label{sec:eval-method}

Earlier, we mentioned that we selected $57$ tasks spanning $26$ reasoning skills from \supernishort to evaluate our finetuned models. To meet our criteria, as detailed in Section \ref{sec:niv2-tasks}, each task had to fulfill two conditions. The second condition required that the task can be considered a classification task. That means there is a discrete set of candidates (one of which is correct) and thereby treating it as a classification problem where the highest-scoring candidate is considered the answer. To ensure this, we utilized a straightforward heuristic: we only sampled tasks that had no more than $10$ possible candidate answers. 

\paragraph{Classification Method} To determine the correct answer, we conduct a forward pass for each potential candidate answer and utilize a scoring function to measure the likelihood that the candidate tokens follows the input, similar to \citet{gpt3}. This process is repeated four times using distinct scoring functions, as detailed in the subsequent paragraph. The highest accuracy score from the four scoring functions is considered as result of the task. 

\paragraph{Scoring Functions} This is considered the fourth dimension of this work since we evaluate each task using four different scoring functions and take the maximum accuracy as the result. The four scoring functions used are as follows: (1) \textbf{mean}, which involves computing the average of the log probabilities of candidate tokens, also referred to as token score. (2) \textbf{unconditional-norm}, which computes the difference between the sum of token scores of the candidate when unconditioned by any previous tokens and the sum of candidate token scores when conditioned by previous input. (3) \textbf{suffix}, which computes the sum of the conditioned candidate's token scores alone. Finally, (4) \textbf{sum}, which involves calculating the sum of all the token scores passed to the model. The reason we employed different functions is that we observed significant gains in performance when using one scoring function over the other for specific tasks. Therefore, in order to ensure fairness across all tasks, we selected the highest accuracy over all scoring functions for each task.

\section{Results \& Findings}

In this section, we present the results and findings of our experiments. First, we illustrate in Figure \ref{fig:result_main} the outcome of our evaluation on the effectiveness of finetuned models as compared to the vanilla OPT model, across three different scales when using both fewshot prompting with and without explanations. Furthermore, we observe a monotonic increase in the performance of each model as we increase the scale under those two prompting condition, which indicates a positive correlation between the model's capacity and its overall performance. However, we note that this trend does not apply to the zeroshot prompting method, since we are testing out-of-distribution tasks and that the finetuned models were trained with fewshot exemplars in their context. This leads us to focus only on the fewshot prompting methods, with and without explanations, for the remaining of our evaluations. Specifically, we investigate the impact of finetuning the OPT models on reasoning datasets, as compared to the vanilla OPT model, and explore the effect of explanations during finetuning and prompting, both in terms of the reasoning skill.


\subsection{Model Performance for Reasoning Skills}
\label{sec:model-perf-skills}

The results reported in this and the following section are the classification accuracy of each reasoning skill across different conditions, such as model sizes and fewshot prompting methods. Table \ref{tab:results-opt-r-sig} shows the reasoning skills where either \ourmodel{RE} or \ourmodel{R} are significantly better than the vanilla OPT model, as measured by Welch's t-test, where $p < 0.05$. Conversely, Table \ref{tab:results-opt-sig} show the reasoning skills where the vanilla OPT model performs significantly better than either of its finetuned counterparts.

\begin{table}[ht!]
\centering
\begin{tabular}{lccc}
\toprule
\textbf{Skill} &     \textbf{OPT} &     \textbf{\ourmodel{R}} &     \textbf{\ourmodel{RE}} \\ 
\midrule
\textbf{Numerical           } &  44.8 &  \textbf{65.2}* &  64.7* \\
\textbf{Analogical          } &  49.0 &  \textbf{62.9}* &  60.8* \\
\textbf{Counting            } &  19.8 &  13.1\phantom{*} &  \textbf{31.3}* \\
\textbf{Physical            } &  38.2 &  37.8\phantom{*} &  \textbf{49.1}* \\
\textbf{Entailment          } &  42.6 &  47.2\phantom{*} &  \textbf{51.6}* \\
\textbf{Social Int    } &  34.1 &  \textbf{43.0}* &  40.1\phantom{*} \\
\textbf{Objects             } &  54.3 &  \textbf{62.6}* &  59.9* \\



\bottomrule
\end{tabular}
\caption{Performance as a function of the reasoning skills where \ourmodel{RE} or \ourmodel{R} performs significantly better than the OPT model as measured by Welch's t-test ($p < 0.05$) denoted by the * symbol. The performance is measured across Fewshot and Fewshot-E prompting, the three different scales and tasks under the corresponding reasoning skill. Best result indicated in \textbf{bold}.}
\label{tab:results-opt-r-sig}
\end{table}

\begin{table}[ht!]
\centering
\begin{tabular}{lccc}
\toprule
\textbf{Skill} &     \textbf{OPT} &     \textbf{\ourmodel{R}} &     \textbf{\ourmodel{RE}} \\ 
\midrule
\textbf{Argument           } &  \textbf{57.9} &  46.1$^-$ &  48.7$^-$ \\
\textbf{TE - Deductive     } &  \textbf{36.0} &  29.0$^-$ &  29.4$^-$ \\
\textbf{Commonsense        } &  \textbf{33.4} &  29.7\phantom{$^-$} &  28.8$^-$ \\


\bottomrule
\end{tabular}
\caption{Performance as a function of the reasoning skill where OPT performs significantly better than either \ourmodel{R} or \ourmodel{RE} as measured by Welch's t-test ($p < 0.05$) denoted by the $^-$ symbol. The performance is measured across Fewshot and Fewshot-E prompting, the three different scales and tasks under the corresponding reasoning skill. TE is Textual Entailment.}
\label{tab:results-opt-sig}
\end{table}

The results reveal that the finetuned variants of the OPT model demonstrate a significant improvement on seven distinct reasoning skills, with particular emphasis on the \textit{Numerical} and \textit{Analogical} reasoning tasks. Specifically, for the \textit{Mathematical Counting} skill, the \ourmodel{RE} variant outperforms both the \ourmodel{R} and OPT models, underscoring the criticality of incorporating explanations during the finetuning process for mathematical datasets. Likewise, the \textit{Physical Reasoning} tasks exhibit a similar trend. On the other hand, we can see that for the \textit{Argument}, \textit{Deductive Textual Entailment} and \textit{Commonsense} skills the non-finetuned version outperforms considerably.

\subsection{Fine-Grained Skill Analysis}
\label{sec:analysis}

\begin{table*}[ht!]
\centering
\begin{tabular}{lcccccc}
\toprule 
{} &                 \multicolumn{2}{c}{\textbf{OPT}} &                \multicolumn{2}{c}{\textbf{\ourmodel{R}}} &                \multicolumn{2}{c}{\textbf{\ourmodel{RE}}}  \\
\textbf{Skill                } & \textbf{Fewshot} &        \textbf{Fewshot-E} &      \textbf{Fewshot} &       \textbf{Fewshot-E} &      \textbf{Fewshot} &       \textbf{Fewshot-E}  \\
\midrule
\textbf{Numerical            } &           \cellcolor[HTML]{ffffff}{39.9} &           \cellcolor[HTML]{b0ecb0}{49.7} &           \cellcolor[HTML]{34cd34}{65.1} &  \cellcolor[HTML]{32cd32}{\textbf{65.3}} &           \cellcolor[HTML]{37ce37}{64.7} &           \cellcolor[HTML]{36ce36}{64.8} \\
\textbf{Analogical           } &           \cellcolor[HTML]{bbeebb}{51.9} &           \cellcolor[HTML]{ffffff}{46.2} &  \cellcolor[HTML]{32cd32}{\textbf{63.3}} &           \cellcolor[HTML]{3ccf3c}{62.5} &           \cellcolor[HTML]{51d551}{60.7} &           \cellcolor[HTML]{4fd44f}{60.9} \\
\textbf{Objects              } &           \cellcolor[HTML]{ffffff}{53.5} &           \cellcolor[HTML]{e9fae9}{55.1} &           \cellcolor[HTML]{94e594}{61.4} &  \cellcolor[HTML]{74dd74}{\textbf{63.8}} &           \cellcolor[HTML]{a7eaa7}{60.0} &           \cellcolor[HTML]{abebab}{59.7} \\
\textbf{Social Interactions  } &           \cellcolor[HTML]{ffffff}{33.6} &           \cellcolor[HTML]{f0fbf0}{34.7} &  \cellcolor[HTML]{76de76}{\textbf{43.8}} &           \cellcolor[HTML]{8ae38a}{42.3} &           \cellcolor[HTML]{a6eaa6}{40.2} &           \cellcolor[HTML]{a9eaa9}{40.0} \\
\textbf{Textual Entailment   } &           \cellcolor[HTML]{eefbee}{43.3} &           \cellcolor[HTML]{ffffff}{42.0} &           \cellcolor[HTML]{baefba}{47.1} &           \cellcolor[HTML]{b8eeb8}{47.3} &  \cellcolor[HTML]{7adf7a}{\textbf{51.9}} &           \cellcolor[HTML]{83e183}{51.2} \\
\textbf{Grammatical          } &           \cellcolor[HTML]{ffffff}{54.4} &           \cellcolor[HTML]{f6fdf6}{55.1} &           \cellcolor[HTML]{a4e9a4}{61.2} &           \cellcolor[HTML]{b4edb4}{60.0} &           \cellcolor[HTML]{9ae79a}{62.0} &  \cellcolor[HTML]{8be38b}{\textbf{63.1}} \\
\textbf{Multihop             } &           \cellcolor[HTML]{beefbe}{36.6} &           \cellcolor[HTML]{ffffff}{31.7} &           \cellcolor[HTML]{9fe79f}{38.9} &  \cellcolor[HTML]{92e492}{\textbf{39.9}} &           \cellcolor[HTML]{97e597}{39.5} &           \cellcolor[HTML]{b9eeb9}{37.0} \\
\textbf{Symbols              } &           \cellcolor[HTML]{ffffff}{44.2} &           \cellcolor[HTML]{d7f5d7}{47.2} &           \cellcolor[HTML]{9be69b}{51.7} &           \cellcolor[HTML]{9ae69a}{51.8} &           \cellcolor[HTML]{99e699}{51.9} &  \cellcolor[HTML]{92e492}{\textbf{52.4}} \\
\textbf{Spatial              } &           \cellcolor[HTML]{ffffff}{44.1} &           \cellcolor[HTML]{d7f5d7}{47.1} &           \cellcolor[HTML]{b3ecb3}{49.8} &  \cellcolor[HTML]{99e699}{\textbf{51.8}} &           \cellcolor[HTML]{b6edb6}{49.6} &           \cellcolor[HTML]{bbeebb}{49.2} \\
\textbf{Social Situations    } &           \cellcolor[HTML]{ffffff}{46.3} &           \cellcolor[HTML]{fbfefb}{46.6} &           \cellcolor[HTML]{a5e9a5}{53.2} &           \cellcolor[HTML]{a5e9a5}{53.2} &           \cellcolor[HTML]{b6edb6}{51.9} &           \cellcolor[HTML]{b1ecb1}{52.3} \\

\midrule

\textbf{Counting             } &           \cellcolor[HTML]{b9eeb9}{19.6} &           \cellcolor[HTML]{b5edb5}{20.0} &           \cellcolor[HTML]{f7fdf7}{13.5} &           \cellcolor[HTML]{ffffff}{12.7} &           \cellcolor[HTML]{51d551}{29.8} &  \cellcolor[HTML]{32cd32}{\textbf{32.9}} \\
\textbf{Physical             } &           \cellcolor[HTML]{ffffff}{35.8} &           \cellcolor[HTML]{bdefbd}{40.6} &           \cellcolor[HTML]{f0fbf0}{36.9} &           \cellcolor[HTML]{d6f5d6}{38.8} &           \cellcolor[HTML]{57d657}{48.1} &  \cellcolor[HTML]{3dd03d}{\textbf{50.0}} \\
\textbf{Logical              } &           \cellcolor[HTML]{ffffff}{31.7} &           \cellcolor[HTML]{e9fae9}{33.4} &           \cellcolor[HTML]{e5f9e5}{33.7} &           \cellcolor[HTML]{dff7df}{34.1} &           \cellcolor[HTML]{bbefbb}{36.9} &  \cellcolor[HTML]{a7eaa7}{\textbf{38.4}} \\

\midrule

\textbf{Temporal             } &  \cellcolor[HTML]{5ad75a}{\textbf{50.7}} &           \cellcolor[HTML]{68da68}{49.7} &           \cellcolor[HTML]{bdefbd}{43.4} &           \cellcolor[HTML]{93e593}{46.5} &           \cellcolor[HTML]{78de78}{48.5} &           \cellcolor[HTML]{ffffff}{38.5} \\
\textbf{Argument             } &           \cellcolor[HTML]{78de78}{55.8} &  \cellcolor[HTML]{3dd03d}{\textbf{60.1}} &           \cellcolor[HTML]{fafefa}{46.3} &           \cellcolor[HTML]{ffffff}{45.9} &           \cellcolor[HTML]{daf6da}{48.6} &           \cellcolor[HTML]{d7f5d7}{48.8} \\
\textbf{TE - Deductive       } &           \cellcolor[HTML]{b1ecb1}{33.7} &  \cellcolor[HTML]{73dd73}{\textbf{38.3}} &           \cellcolor[HTML]{ffffff}{27.9} &           \cellcolor[HTML]{e1f8e1}{30.1} &           \cellcolor[HTML]{f0fbf0}{29.0} &           \cellcolor[HTML]{e4f8e4}{29.9} \\
\textbf{Relational           } &           \cellcolor[HTML]{daf6da}{47.4} &  \cellcolor[HTML]{aaeaaa}{\textbf{51.1}} &           \cellcolor[HTML]{d8f5d8}{47.6} &           \cellcolor[HTML]{d4f4d4}{47.9} &           \cellcolor[HTML]{fcfefc}{44.8} &           \cellcolor[HTML]{ffffff}{44.6} \\
\textbf{Commonsense          } &  \cellcolor[HTML]{aaeaaa}{\textbf{35.0}} &           \cellcolor[HTML]{d4f4d4}{31.8} &           \cellcolor[HTML]{eefbee}{29.8} &           \cellcolor[HTML]{f2fcf2}{29.5} &           \cellcolor[HTML]{ffffff}{28.5} &           \cellcolor[HTML]{f6fdf6}{29.2} \\

\midrule

\textbf{TE - Analogical      } &           \cellcolor[HTML]{ffffff}{16.3} &           \cellcolor[HTML]{e1f8e1}{18.7} &           \cellcolor[HTML]{e2f8e2}{18.6} &  \cellcolor[HTML]{c8f2c8}{\textbf{20.7}} &           \cellcolor[HTML]{e1f8e1}{18.7} &           \cellcolor[HTML]{e8fae8}{18.1} \\
\textbf{Abductive            } &           \cellcolor[HTML]{ffffff}{33.9} &           \cellcolor[HTML]{e5f8e5}{36.1} &  \cellcolor[HTML]{dcf6dc}{\textbf{36.9}} &           \cellcolor[HTML]{f9fef9}{34.4} &           \cellcolor[HTML]{fbfefb}{34.2} &           \cellcolor[HTML]{effbef}{35.3} \\
\textbf{Ethics               } &           \cellcolor[HTML]{f5fdf5}{26.8} &           \cellcolor[HTML]{ffffff}{25.8} &           \cellcolor[HTML]{f8fdf8}{26.5} &           \cellcolor[HTML]{fefffe}{25.9} &           \cellcolor[HTML]{fbfefb}{26.2} &  \cellcolor[HTML]{edfbed}{\textbf{27.6}} \\
\textbf{Deductive            } &           \cellcolor[HTML]{ffffff}{39.4} &           \cellcolor[HTML]{f5fdf5}{40.4} &           \cellcolor[HTML]{ffffff}{39.4} &           \cellcolor[HTML]{f5fdf5}{40.4} &           \cellcolor[HTML]{f9fef9}{40.0} &  \cellcolor[HTML]{eefbee}{\textbf{41.1}} \\
\textbf{Causal               } &           \cellcolor[HTML]{f2fcf2}{50.2} &  \cellcolor[HTML]{eefbee}{\textbf{50.6}} &           \cellcolor[HTML]{fdfffd}{49.1} &           \cellcolor[HTML]{ffffff}{48.9} &           \cellcolor[HTML]{f3fcf3}{50.1} &           \cellcolor[HTML]{effbef}{50.5} \\
\textbf{Scientific           } &           \cellcolor[HTML]{fefffe}{23.4} &           \cellcolor[HTML]{ffffff}{23.3} &           \cellcolor[HTML]{f5fdf5}{24.3} &           \cellcolor[HTML]{f3fcf3}{24.5} &  \cellcolor[HTML]{eefbee}{\textbf{25.0}} &           \cellcolor[HTML]{f3fcf3}{24.5} \\
\textbf{Numerical Commonsense} &  \cellcolor[HTML]{ffffff}{\textbf{59.5}} &           \cellcolor[HTML]{ffffff}{59.2} &           \cellcolor[HTML]{ffffff}{59.0} &           \cellcolor[HTML]{ffffff}{59.0} &           \cellcolor[HTML]{ffffff}{59.2} &           \cellcolor[HTML]{ffffff}{59.4} \\
\textbf{Strings              } &           \cellcolor[HTML]{ffffff}{60.7} &           \cellcolor[HTML]{ffffff}{60.7} &           \cellcolor[HTML]{ffffff}{61.1} &  \cellcolor[HTML]{ffffff}{\textbf{61.2}} &           \cellcolor[HTML]{ffffff}{60.7} &           \cellcolor[HTML]{ffffff}{60.7} \\
\bottomrule
\end{tabular}
\caption{Classification accuracy results achieved by different models as a function of the reasoning skill and few-shot prompting method employed. The best accuracy obtained for each reasoning skill is highlighted in bold. The cells are shaded with colors ranging from green to white to indicate their position in the accuracy spectrum. Reasoning skills with smaller variance in achieved results are assigned a lighter shade of green to convey the extent of similarity between models. The first block highlights skills where the finetuned models perform notably better than the vanilla OPT. The second block emphasizes the skills where \ourmodel{RE} outperforms other models. In contrast, the third block showcases the skills where OPT outperforms the other models. Lastly, the fourth block identifies skills where the choice of model or prompting method has little impact on the overall performance.}
\label{tab:results-color}
\end{table*}

Table \ref{tab:results-color} shows the classification accuracy results obtained from the three models, in relation to the reasoning skill and few-shot prompting method used. The best accuracy value for each reasoning skill is indicated in \textbf{bold}, and the cells are shaded with colors ranging from green to white to indicate their position in the accuracy spectrum of each reasoning skill. The skills with similar performance across different models are assigned a lighter shade of green, indicating that their color spectrum ends earlier than that of other skills where the difference in performance between models is more significant. The table is divided into four blocks to distinguish effects of finetuning and prompting methods on reasoning skills: the first block showcases skills where the finetuned (\ourmodel{RE} and \ourmodel{R}) models outperform the vanilla OPT model, the second block highlights skills where \ourmodel{RE} has better accuracy than other models therefore illustrating the importance of finetuning on explanations on those skills. The third block displays skills where OPT outperforms other models showing that finetuning actually hurts performance in this case, and the fourth block identifies skills where the choice of model or prompting method has little impact on the overall performance.

\paragraph{Explanations' Effect} 

One of the central questions that we sought to investigate in this study is the extent to which explanations play a role in improving the reasoning capabilities of OPT models during finetuning and prompting. The results presented in Table \ref{tab:results-std-prompt} suggest that the presence or absence of explanations in the fewshot examples employed for prompting does not \textit{significantly} impact the performance of the model when the model is finetuned on reasoning datasets. Concretely, in Table \ref{tab:results-std-prompt}, we present the variance of the absolute accuracy difference for each model across reasoning skills by excluding the \textit{Temporal} skill, which was identified as an outlier. Specifically, we compute the difference between the two corresponding columns for each model in Table \ref{tab:results-color}. These values provide insights into the impact of including explanations during prompting on the performance of the models. Our findings reveal that the difference is negligible for \ourmodel{R} and \ourmodel{RE} models, suggesting that the choice of prompting method does not significantly affect the model's accuracy. However, for the vanilla OPT model, the difference is more substantial, emphasizing the importance of employing explanations during fewshot prompting. However, the mean performance of each model across the distinct fewshot prompting methods demonstrates a slight yet consistent increase in classification accuracy, from Fewshot to Fewshot-E (incorporating explanations), as well as from OPT to \ourmodel{R} to \ourmodel{RE} models showing that explanations do have a small effect on performance during both finetuning and prompting.

\begin{table}[ht!]
\centering
\begin{tabular}{lccc}
\toprule
\textbf{Model} &     \textbf{Std(|F-FE|)} & \textbf{Avg(F)} & \textbf{Avg(FE)}  \\
\midrule
\textbf{OPT    } &  2.31  & 40.68 & 41.82\\
\textbf{\ourmodel{R}  } &  0.84 & 43.44 & 43.68 \\
\textbf{\ourmodel{RE} } &  0.78  & 44.49 & 44.86 \\
\bottomrule
\end{tabular}
\caption{The first column shows the variance of the absolute difference in accuracy for each model across different reasoning skills, when using Fewshot (F) and Fewshot-E (FE) prompting methods. The second and third columns show the average performance of each model across each prompting method. Results are obtained after dropping the outlier \textit{Temporal} skill.}
\label{tab:results-std-prompt}
\end{table}

\section{Related Work}
\label{sec:related}

\paragraph{Reasoning LLMs} 
LLMs have made significant advancements in the field of NLP and related areas \cite{gpt3, palm2022, chung2022scaling}, especially with the advent of the pre-train, prompt, and predict paradigm \cite{Liu2021PretrainPA}. This paradigm has enabled these models to solve a multitude of tasks through in-context fewshot or zeroshot learning using instructions \cite{flan, iyer2022opt}. However, their reasoning abilities have been a subject of debate in recent literature \cite{Huang2022TowardsRI, AlKhamissi2022ARO}. Several studies suggest that increasing the size of an LM trained through the same next-token prediction method can lead to the emergence of complex behaviors \cite{Wei2022EmergentAO}, including reasoning. For instance, some research has demonstrated that sufficiently large LMs can use chain-of-thought prompting \cite{wei2022chain} to simulate human-like reasoning. Other studies have shown that the addition of a simple prompt, such as "Let's think step-by-step" \cite{kojima2022large} can elicit reasoning abilities in LLMs by generating explicit reasoning steps before decoding the final answer. However, some researchers contend that emulating the human reasoning thought process is distinct from claiming that the model can truly reason \cite{wei2022chain}. 

\paragraph{Finetuned LLMs} 

Concurrent studies have finetuned LLMs to follow instructions to improve their generalization ability to unseen tasks through zero and fewshot learning \cite{iyer2022opt, chung2022scaling}. However, our approach differs in that we only finetune on a selected number of open-source datasets that provide explanations for each instance. This enables us to focus on the importance of explanations during finetuning in the context of reasoning skills. While concurrent works, such as \cite{iyer2022opt, niv2}, have experimented with different prompting methods during finetuning and inference, our study focuses primarily on evaluating the reasoning ability of the finetuned models across a set of reasoning skills. Other concurrent studies have explored the impact of finetuning on a set of held-out reasoning tasks \cite{Yu2022ALERTAL}, but their evaluation approach, which involves generating answers, may be influenced by various factors such as decoding strategy, decoding parameters, and prompt templates. In contrast, we adopt a rank classification approach similar to \cite{gpt3}, which better captures the reasoning performance of the model being evaluated, in addition to covering a larger number of reasoning skills and tasks.

\section{Conclusion}

In this study, we investigated the impact of incorporating explanations during finetuning and prompting on three different sizes of the OPT model. Through a systematic and comprehensive evaluation process that considered three key dimensions, we found that while explanations did provide a small improvement in performance, the effect was not significant when incorporated in the in-context demonstrations during inference for the finetuned models. Additionally, our results showed that both finetuned models exhibited significant improvements in reasoning skills such as \textit{Numerical}, \textit{Analogical} and \textit{Reasoning on Objects}. Moreover, we demonstrated that skills such as \textit{Physical}, \textit{Counting}, and \textit{Textual Entailment} benefited from incorporating explanations during the finetuning process. Overall, our findings provide insights into the impact of incorporating explanations on the reasoning capabilities of LLMs and offer guidance on which reasoning skills would benefit most from the inclusion and exclusion of explanations during finetuning and prompting.





\section*{Limitations}

While our study provides valuable insights into the impact of finetuning on reasoning performance and the role of explanations during finetuning and prompting with respect to various reasoning skills, there are several limitations to our work. Firstly, we only consider a single LLM, OPT, as our base model. Our results may not generalize to other LLMs with different architectures or pretraining objectives. Secondly, we only use a limited set of reasoning datasets for finetuning due to the limited availability of open-source datasets with explanations. However, it is possible that our findings may not hold for models finetuned on larger closed datasets as usually seen in real-world scenarios. Thirdly, our experiments only cover a limited range of model sizes due to limitations in computational budget, therefore it is possible that our findings may not hold for much larger models. Finally, we only consider finetuning using fewshot prompting conditions in our experiments, and it is possible that our findings may not hold for models finetuned without in-context exemplars. Overall, while our study provides valuable insights into the impact of finetuning and explanations on reasoning performance, further research is needed to investigate these factors across a broader range of models, datasets, and finetuning strategies.

\section*{Ethics Statement}

This work is based on analyzing and evaluating the performance of LLMs on reasoning tasks using existing public datasets. No personally identifiable information or sensitive data was collected or used in this research. We acknowledge the potential risks of developing LLMs, including their potential impact on spreading misinformation, generating unwanted content and the exacerbation of existing biases in datasets. Our work aims to contribute to improving the transparency and understanding of how LLMs can be optimized for specific reasoning skills. We hope our findings will inspire further research on developing ethical and responsible approaches for developing and deploying LLMs.

\bibliography{custom}
\bibliographystyle{acl_natbib}

\appendix

\begin{table*}[ht!]
\centering
\begin{tabular}{@{}p{0.1\linewidth}p{0.7\linewidth}p{0.2\linewidth}@{}}
\toprule
\textbf{Dataset} & \textbf{Task Definition}  & \makecell[r]{\textbf{Options}}        \\ \midrule
AQuA         & \makecell[l]{You are given an algebraic word question. Questions in this task often \\ requires executing a series of arithmetic operations to obtain a final answer.\\ You are also given 5 answer options (associated with 'A', 'B, 'C', 'D', 'E').\\ Do not generate anything else apart from one of the following characters:\\ "A", "B", "C", "D", "E" and the corresponding explanation.} & \makecell[r]{-A \\ -B \\ -C \\ -D \\ -E} \\ \\
CoQA             & You are given a passage that contains a conversation and a question. The task is to answer the question and provide an explanation that highlights the corresponding evidence in the passage.                                                                                                                                                                            &  \makecell[r]{Free-form text}                     \\  \\
CoS-E            & \makecell[l]{You are given a passage that contains a sentence and a question. The task \\ is to answer the question by selecting one of the provided choices.} &  \makecell[r]{Select one of \\ the provided choices}                       \\ \\
ECQA             & \makecell[l]{You are given a question that requires commonsense reasoning. The task \\ is to answer the question by selecting one of the provided choices.}  &  \makecell[r]{Select one of \\ the provided choices}                     \\ \\
ESNLI            & \makecell[l]{You will be presented with a premise and a hypothesis sentence. The \\ task is to determine whether the hypothesis sentence entails (implies),\\ contradicts (opposes), or is neutral with respect to the given premise \\ sentence. Please answer with "Contradiction", "Neutral",or "Entailment".} &  \makecell[r]{-Contradiction \\ -Neutral \\ -Entailment}                     \\ \\
GSM8K            & You will be presented with a passage that contains a grade school math word problem. The task is to answer the question by performing a series of arithmetic operations to obtain a final answer.                                                                                                                                                                        &  \makecell[r]{Number}                       \\ \\
ProofWriter      & \makecell[l]{You are given a sequence of facts and rules followed by a question. The \\ task is to answer the question using "Yes", "No" or "Unknown".}                                                                                                                                                                                                                                   &  \makecell[r]{ -Yes \\ -No \\ -Unknown}                      \\ \\
StrategyQA       & \makecell[l]{You are given a sentence and a question. The required reasoning steps are \\ implicit in the question. The task is to answer the question using "Yes" or \\ "No" then provide a strategy that explains the answer by decomposing it \\ into a number of steps.}                                                                                                                    &  \makecell[r]{ -Yes \\ -No}                        \\ \bottomrule
\end{tabular}%
\caption{Task definition and options used for each of the finetuning reasoning datasets.}
\label{tab:finetuning-datasets}
\end{table*}
\newpage

\section{Finetuning Task Definition and Options}
\label{app:finetuning-ds}

Table \ref{tab:finetuning-datasets} shows the task definition and options provided as input to the template shown in Figure \ref{fig:template} during finetuning the OPT models on the reasoning datasets.

\end{document}